\documentclass[10pt,twocolumn,letterpaper]{article}

\usepackage{cvpr}
\usepackage{times}
\usepackage{epsfig}
\usepackage{graphicx}
\usepackage{amsmath}
\usepackage{amssymb}
\usepackage[utf8]{inputenc}

\usepackage[breaklinks=true,bookmarks=false]{hyperref}

\cvprfinalcopy 

\def\cvprPaperID{****} 

\begin{document}

\title{Automated Focal Loss for Image based Object Detection}



\author{Michael Weber\thanks{\tt\small michael.weber@fzi.de}
\qquad Michael Fürst\thanks{{\tt\small David\_Michael.Fuerst@dfki.de} Michael Fürst is now with the German Research Center for Artificial Intelligence }
\qquad J. Marius Zöllner\thanks{{\tt\small zoellner@fzi.de}}
\\FZI Research Center for Information Technology%
}
%


\maketitle

\begin{abstract}
Current state-of-the-art object detection algorithms still suffer the problem of imbalanced distribution of training data over object classes and background. Recent work introduced a new loss function called focal loss to mitigate this problem, but at the cost of an additional hyperparameter. Manually tuning this hyperparameter for each training task is highly time-consuming.

With automated focal loss we introduce a new loss function which substitutes this hyperparameter by a parameter that is automatically adapted during the training progress and controls the amount of focusing on hard training examples. We show on the COCO benchmark that this leads to an up to 30~\% faster training convergence.
We further introduced a focal regression loss which on the more challenging task of 3D vehicle detection outperforms other loss functions by up to $1.8$ AOS and can be used as a value range independent metric for regression.
\end{abstract}

\section{Introduction}

State-of-the-art object detection based on Convolutional Neural Networks~(CNNs) currently can be seen as a competition between so called one-stage detectors~\cite{liu16} and two-stage approaches~\cite{he17}. While the latter achieves better accuracy performances e.~g. on one of the currently challenging object detection benchmarks, the COCO benchmark~\cite{lin14}, they usually suffer from longer run times. As the number of candidate objects processed in the second stage is not known in advance, the run times furthermore are only roughly predictable in advance. 

In contrast, one-stage detectors usually have faster and predictable run times, but are suffering from worse detection accuracy. In~\cite{lin18} the class imbalance has been identified as a source for this performance gap. Two-stage approaches avoid the imbalance problem as they filter most of the background before classification due to their first region proposal step (\eg RPN~\cite{ren15}, Selective Search~\cite{uijlings13}, Edge Boxes~\cite{zitnick14} or DeepMask~\cite{pinheiro15, pinheiro16}). In the second stage, methods like fixed background ratio~\cite{girshick14} or online hard example mining~(OHEM)~\cite{shrivastava16} are used, to finally balance the training data.

\begin{figure}[t]
\centering
\includegraphics[width=0.46\textwidth]{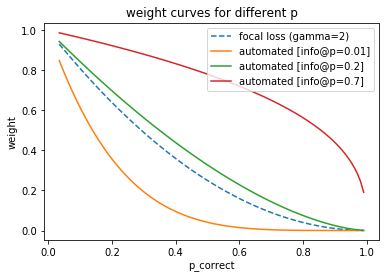}
\caption{Focal loss has introduced a method of weighing samples based on their prediction quality. However this weighing is constant during training (blue dashed line). We propose to automatically change the focus dependent on the training progress modeled as the expected probability of a correct prediction $\hat{p}_{correct}$. When the progress is low the focus is only on the worst predictions ($\hat{p}_{correct} = 0.01$, orange). As the progress improves (green) the automated loss shifts the focus to include better predictions only down weighting excellent predictions ($\hat{p}_{correct} = 0.7$, red).}
\label{img:title}
\end{figure}

For one-stage detectors, the situation is more challenging: As the detector needs to learn to distinguish between foreground classes and all possible background scenery it has to \emph{see} a lot of possible background data. For this reason, the mentioned two-stage methods are not easily applicable to one-stage detectors. 
As a lever to increase accuracy the loss functions receive rising spotlight.
Early one-stage object detectors made use of static loss functions -- meaning that the hyperparameters of the loss were not changed during training of the network. As a first step to mitigate the class imbalance problem, so called $\alpha$-balancing was added to the loss function to weight the losses for different classes/background according to their relative frequency. This mechanism was introduced to prevent the detectors from always predicting the dominant class which usually is background. A first step towards a dynamic loss function was done by applying $\alpha$-balancing to each minibatch separately.

With RetinaNet \cite{lin18}, additionally to the class distribution dynamics, a data centric dynamic was added to the loss function they called \textit{focal loss}. For each prediction, its difficulty in each iteration is calculated based on the estimated probability for the correct class. The influence of this information to the loss function is controlled by adding a manually tuned hyperparameter $\gamma$.

\begin{figure}[t]
\centering
\includegraphics[width=0.46\textwidth]{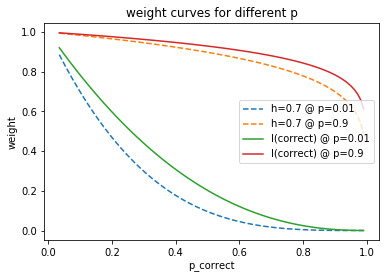}
\caption{We propose two methods to estimate $\gamma$ dependent on the training progress. The first method controls the relative weight mass assigned to examples that perform below average. Having this at a percentage at 0.7 percent between $p$ and 1.0 performed best in our experiments. The second method uses the Shannon information of a correct prediction. $I(correct)$. The weight distribution they assign to a given training progress is similar. With a low training progress both assign weights only to the worst examples (bottom left) while at a high training progress both only reduce the weights for the best predicted examples (top right).}
\label{img:info_vs_h}
\end{figure}

In this paper, we propose a new loss function based on the focal loss~\cite{lin18} to automatically handle the class imbalance problem. 
Our \emph{Automated Focal Loss} no longer relies on the manually tuned hyperparameter $\gamma$ for balancing between easy and hard training examples. We present two methods for automatically adapting the dynamic difficulty term during training process as shown in Figures~\ref{img:title} and \ref{img:info_vs_h}.

Finally we show that with our automated focal loss training converges to the same AP as the static focal loss~\cite{lin18} on the COCO benchmark, while converging in 30~\% less time.
Furthermore we tested our automated focal regression loss on the challenging \emph{KITTI 3D Object Detection} dataset~\cite{geiger12} demonstrating its effectiveness on less extensive datasets.
On this dataset we outperform other losses and show that adding automated focal regression adds a slight edge to our loss over only using automated focal classification.

\section{Related Work}

\textbf{CNN based Object Detection:}
The idea of detecting objects with the help of CNNs goes back to 1998 when LeCun et al. \cite{lecun98} extended the initial CNN classification approach also done by LeCun et al. \cite{lecun89} to the ability of detecting objects within an image. This network transformation which later was named Fully Convolutional Network (FCN)~\cite{long15} implicitly models a sliding window within the network structure.

This architecture was the foundation of OverFeat~\cite{sermanet13}, one of the first modern CNN based object detectors. Together with SSD~\cite{liu16, fu17}, YOLO~\cite{redmon16, redmon17, redmon18} and the most recent RetinaNet~\cite{lin18} OverFeat spans the class of so called one-stage object detectors.
They consist of one single convolutional network which extracts features, predicts object positions and classifies the objects.

The other large class of CNN object detectors are the so called two-stage approaches which adopted the two stages of classical object detection approaches into CNN architectures. 
In this classical chain, detection and classification of objects are separated into two different steps where classification between foreground object and background is applied upon the detection results. 
Detections in these approaches are usually called region proposals and do have to be thought of as candidate regions within an image where an object could be present.
Region-CNN \cite{girshick14} was the first of these approaches while still using a Support Vector Machine (SVM) for the classification task after extracting features with a CNN. 
The following adaptions of this approach~\cite{girshick15, ren15, he15, he17, lin17b} turned the classification part completely into a CNN and enhanced the region proposal generation which dramatically increased the detection speed.

The differences between the one-stage and two-stage approaches have been mainly in execution speed and detection accuracy. While two-stage approaches usually reached a higher detection accuracy but to the cost of a longer execution run time. With the emergence of RetinaNet~\cite{lin18} using a focal loss function with an one-stage approach, they could achieve an comparable accuracy to recent two-stage approaches.

\textbf{3D Vehicle Detection:}
Detection is not only limited to the case of two dimensional detection in the image plane.
Recent approaches~\cite{poirson16, kehl17} have extended the approach of two dimensional detection to the challenging task of detection in the space of the three dimensional world.
The subfield of three dimensional vehicle detection is actively researched due to the development of autonomous vehicles.
There are various approaches to estimate the pose of vehicles.
They can be broadly divided into two categories.
The first category indirectly estimates the pose and dimensions of the vehicle in the three dimensional space.
In general these approaches estimate features or constraints onto which a template is fit.
Approaches like DeepMANTA~\cite{chabot17}, BoxCars~\cite{sochor18, mousavian17}  and MergeBox~\cite{gahlert18} are some of the best performing in this category.
On the other hand there is the category of networks which directly predict the bounding box in three dimensional space like~\cite{chen17} and the simple 3D detector we propose to test our loss on the challenging task of three dimensional detection.
The latter category does not have issues with the approximation due to limitations of the template fitting and is a solution only requiring neural networks and no complex custom made post processing logic.

\textbf{Loss Functions for Object Detection:}
The loss function as one crucial part of each one-stage object detection system usually can be divided into three parts: The loss function used for object classification, the analogue function for object position regression as well as the method for combining these two functions into the final loss function. For classification loss usually Softmax cross entropy~\cite{sermanet13, redmon17, fu17} or Sigmoid cross entropy~\cite{lin18} are used. In \cite{redmon16}, a L2 loss is used for the classification part. So called $\alpha$-balancing is used to balance the influence of different samples based on their class distribution in the training set \cite{liu16}. In \cite{liu16} also a dynamic weighting of training samples according to their difficulty is done.

For the position regression task, the L2 loss~\cite{sermanet13, redmon16}, its variant smooth L2 loss~\cite{liu16} or the similar smooth L1 loss~\cite{fu17} are commonly used functions.
The combination of different loss functions is better known from multitask networks like Multinet~\cite{teichmann18} or \cite{kendall17} but classification and detection can also be seen as tasks within a multitask network. Object detection networks usually make use of a simple sum of losses~\cite{redmon16} or a weighted sum of losses~\cite{sermanet13, liu16, fu17}.

\section{Automated Focal Loss}

The focal loss as introduced by~\cite{lin18} eliminates the need for hard negative mining~\cite{shrivastava16} and helps with the problem of imbalanced data.
However, as stated in the original work, focal loss still needs $\alpha$-balancing to successfully achieve competitive results on the COCO dataset~\cite{lin14}.
By automating the focal loss and dynamically adjusting the focus ($\gamma$) to the current training progress, we found that neither hard negative mining nor $\alpha$-balancing or any other method of simplifying the training is required.
Automated focal loss enables the neural network to automatically focus itself onto the most important examples for the current training progress.

To correctly define and apply automated focal loss, a formal definition as done by~\cite{lin18} is required.
The first step is to define the probability of the correct class.
This means assigning the probability mass of the class that is the correct solution for the task to a variable.
In case of a positive sample, this is the probability that the network assigned the class and in the other case it is the probability that the network did not assign to the class.
This leads to a definition of $p_{correct}$ and the simplified cross entropy for a single \linebreak example as follows:

\begin{equation}
    p_{correct} = \begin{cases}
        p &\text{if $y = 1$,}\\
        1-p &\text{else.}
    \end{cases}
\end{equation}
\begin{equation}
    L_{CE} = - \log(p_{correct})
\end{equation}

The loss can be weighted with a dynamic factor $w$ which is dependent on the probability of the correct result $p_{correct}$.
This factor defines which training samples the network is most focused on at any given point during the training.
\begin{align}
\label{eq:focal_loss}
    L_{focused} &= w \cdot L, &w = (1-p_{correct})^\gamma
\end{align}

So the focused loss as introduced by~\cite{lin18} is dependent on the weight which is computed using $p_{correct}$ and a $\gamma$ which defines the amount of focus that is intended.
Inspecting the derivative for the network variables $\texttt{net}$ reveals the weight $w$ remains unchanged in the derivative as a constant.
\begin{align}
    \frac{ \partial L_{focused} }{\partial \texttt{net}} &= w \cdot \frac{ \partial L }{\partial \texttt{net}}
\end{align}
A high $\gamma$ (\eg $\gamma = 5$) weights down all samples and gradients that are approximately correct while focusing on the examples which are yielding poor performance.
However, as the training progresses the number of poorly classified examples decreases.
This leads to down-weighting an increasing number of samples in the loss and therefore in the gradients as well, leading to a significant drop in training speed due to decreasing gradients.
On the other hand having a too low $\gamma$ (\eg $\gamma = 0$) reduces the impact of the weighting factor $w$.
However, on complex tasks such as COCO and 3D detection, this will render the network incapable of learning the problem.
Especially when the data is imbalanced the network can easily be overwhelmed by the dominant data.

We propose to adapt $\gamma$ during the training progress over time overcoming the limitations of choosing a fixed $\gamma$.
At the beginning of the training $\gamma$ should be high to achieve a good focus on only the poorly predicted training samples.
During training $\gamma$ needs to shift towards 0 to avoid diminishing gradients due to down weighting of well predicted examples.
We decided to model the training progress dependent on the expected probability of the correct prediction $\hat{p}_{correct}$ since this has a direct influence on the expected focal weight (equation~\ref{eq:expected_weight}).
\begin{equation}
\label{eq:expected_weight}
    E \{ \log_\gamma(w) \} = E \{ 1 - p_{correct} \} = 1 - \hat{p}_{correct}
\end{equation}
The expected probability of the correct prediction $\hat{p}_{correct}$ can be approximated by computing the mean over $p_{correct}$ for a training batch.
In case of a small training batch applying a low pass filter like exponential smoothing is recommended. A smoothing via $\hat{p}_{correct} = 0.95 \cdot \texttt{old} + 0.05 \cdot \texttt{new}$ worked best in our experiments.

\subsection{Choosing $\gamma$ dependent on $\hat{p}_{correct}$}

We generally see two options to define $\gamma$ subject to the training progress dependent on $\hat{p}_{correct}$.
The first option is inspired by the observation that the original focal loss has the issue of the expected weight diminishing as the training progresses.
A formal definition to alleviate this issue is to force the integral of weight below the expected probability of the correct class $\hat{p}_{correct}$ to be equal to a fraction $k$ of the total integral of the weight.
\begin{equation}
    k = \frac{\int_{0}^{\hat{p}_{correct}} (1-p)^\gamma dp}{\int_{0}^{1} (1-p)^\gamma dp}
\end{equation}

This equation can be solved for $\gamma$ by integrating and reordering the equation, leading to a formal definition of $\gamma$ dependent on the chosen fraction of the weight that should be assigned to samples with a probability less than the expected probability.
\begin{equation}
    \label{eq:focal_loss_gamma_via_integral}
    \gamma = \frac{\log(1-k)}{\log(1-\hat{p}_{correct})} - 1
\end{equation}

Having defined $\gamma$ this leads to the question what fraction $k$ should be assigned to the examples performing worse than the average.
The focal loss can only focus on poor predictions, if $\gamma$ has a positive value.
Equation~\ref{eq:focal_loss_gamma_via_integral} only yields positive and valid values when $\hat{p}_{correct} < k < 1$.
Having a $k$ close to $\hat{p}_{correct}$ is undesirable, since this will lead to disabling the focal loss after only a few training epochs as $\hat{p}_{correct}$ increases.
Whereas $k = 1$ means disabling the focal loss.
Therefore adapting $k$ dependent on the training progress seems reasonable.
A simple way is to keep $k$ at a certain point between the lower boundary $\hat{p}_{correct}$ and the upper boundary of $1$.
This leads to a definition of $k$ using an interpolation parameter $h$ between the lower and the upper boundary.
\begin{equation}
    k = (h \cdot \hat{p}_{correct} + (1-h) \cdot 1)
\end{equation}
Resulting in a final equation to compute $\gamma$ given $h$ and $\hat{p}_{correct}$.
\begin{equation}
\label{eq:focal_loss_gamma_via_integral_h}
    \gamma = \frac{\log(1- (h \cdot \hat{p}_{correct} + (1-h) \cdot 1) )}{\log(1-\hat{p}_{correct})} - 1
\end{equation}

\begin{figure}[tb]
\centering
\includegraphics[width=0.46\textwidth]{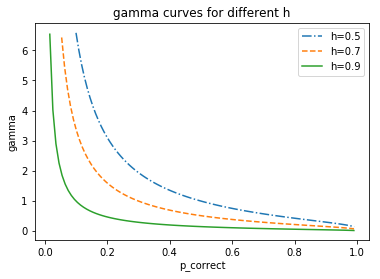}
\caption{Gamma is dependent on $\hat{p}_{correct}$. The parameter $h$ defines the point between $\hat{p}_{correct}$ and $1.0$ that defines the quantile of weight that is assigned to values less than $\hat{p}_{correct}$ (see Equation~\ref{eq:focal_loss_gamma_via_integral_h}). A high value like $0.9$ leads to very low gamma values across the board leading to a low focus, while low values result in higher gamma values and a stronger focus. We found an interpolation value of $h = 0.7$ to be a good specification.}
\label{fig:gamma_h}
\end{figure}

While this definition yields a numerically stable computation of $\gamma$ and fulfills the requirements to model $\gamma$ in an appropriate way.
Namely, for $\hat{p}_{correct} \rightarrow 0$ it diverges to $+\infty$ and for $\hat{p}_{correct} \rightarrow 1$ it converges to $0$ as can be seen in Figure~\ref{fig:gamma_h}.
Even though this approach fulfills the requirement of an automatic focus, it trades the hyperparameter $\gamma$ for the static focus for a hyperparameter $h$.
The hyperparameter gives the flexibility to adapt the loss to the needs of the situation, however if it is not mandatory for success, fewer hyperparameters is the preferred solution.

Defining $\gamma$ inspired by the information as defined by Shannon \cite{shannon48} can eliminate the need for a hyperparameter.
For the loss computation of a neural network, the cross entropy, which has its origin in the information theory by Shannon, is commonly accepted.
Therefore picking the information in a correct prediction of the network $I(correct)$ seems reasonable.
The information has the property that for an optimal network it will converge towards 0.
Since there is no surprise in the outcome, it perfectly matches the ground truth.
On the other hand for a poorly trained network, the information increases without any upper limit.
Defining $\gamma$ as the Shannon information of a correct predictions of the network leads to the following equation:
\begin{equation}
    \label{eq:focal_loss_gamma_via_log}
    \gamma = I(correct) = -log(\hat{p}_{correct})
\end{equation}

\begin{figure}[tb]
\centering
\includegraphics[width=0.46\textwidth]{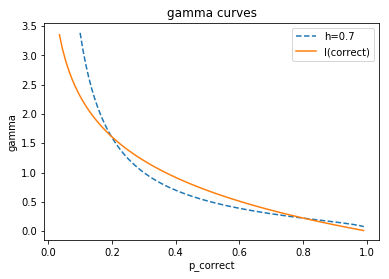}
\caption{Changing $\gamma$ according to the progress of the network can also be modeled as the information in a correct prediction $I(correct) = -\log(\hat{p}_{correct})$. While the curves for an interpolation value of $h=0.7$ and the information $I(correct)$ based approach look similar, in our experiments using the information outperformed using a quantile dependent on $\hat{p}_{correct}$. Furthermore the information based approach has no hyperparameters that need tuning.}
\label{img:gamma_info}
\end{figure}

When checking the definition for sanity the results are as intended.
For a low $\hat{p}_{correct}$ representing a low training progress $\gamma$ is large and therefore the focus on the wrong samples high.
As the training progresses and the $\hat{p}_{correct}$ increases the value of $\gamma$ decreases leading to less focus on the poorly predicted samples (see Figure~\ref{img:gamma_info}).
An ideal network of $\hat{p}_{correct} = 1$ would lead to no focus on poorly predicted examples since there are none.

\subsection{Single Target Classification}

The application of the automated focal loss to a problem requires defining the loss $L$ that should be focused and defining a computation policy for $\hat{p}_{correct}$.
For single target classification the loss $L$ in equation~\ref{eq:focal_loss} should be normal cross entropy loss $L_{CE}$.
Computation of $\gamma$ using $\hat{p}_{correct}$ can be done via equation~\ref{eq:focal_loss_gamma_via_integral_h} or~\ref{eq:focal_loss_gamma_via_log}.
However, the latter should be preferred, since it yields better results in our experiments and needs no hyperparameter.
The required $\hat{p}_{correct}$ can be computed as the average of $p_{correct}$ for the current training batch.
To improve the approximation of $\hat{p}_{correct}$ we found that applying an exponential smoothing worked best.

\subsection{Multi-Target Classification}

In case of multi-target classification, the loss $L$ in equation~\ref{eq:focal_loss} is the binary cross entropy loss $L_{binary\_CE}$.
Computation of $\gamma$ using $\hat{p}_{correct}$ can be done via equation~\ref{eq:focal_loss_gamma_via_integral_h} or~\ref{eq:focal_loss_gamma_via_log}.
Estimating $\hat{p}_{correct}$ is a little more involved than for the case of single target classification.
The probability of a correct class $p_{correct}$ is defined as $p$ in case of a class being active and as $1-p$ when a class is inactive.
The best way to estimate $\hat{p}_{correct}$ depends on the problem that is to be solved and is not as easy as simply averaging $p_{correct}$.
The average of $p_{correct}$ would be dominated by negative classes since typically in a multi-target classification task only a few classes are active at the same time for a single example.

When positive and negative examples are distributed more or less equally, the solution is defining $\hat{p}_{correct}$ as the mean of the probabilities $p$ assigned to all active classes of the example.
Thus avoiding the positive case being overwhelmed by inactive classes.
When an example has no positive classes it is called a negative example.
For negative examples the probabilities not assigned to the classes $1-p$ are averaged.
However, in the case of a task with a lot of negative examples and only few examples with active classes the computation of $\hat{p}_{correct}$ can be easily overwhelmed by the number of negative examples.
Therefore only examples with at least one active class should be considered for computing the average to estimate $\hat{p}_{correct}$.

\subsection{Regression}

\begin{table*}[t]
\centering
3D Detection Decoders\\
\vspace{4pt}
\begin{tabular}{l*{6}{c}r}
\hline
Output            & Value Range & layer type & \#Filters & Filter Size & stride & Activation \\
\hline
probs             & $[0, 1]$ & conv & \#classes & 1x1 & 1,1 & softmax \\
$c_x, c_y$        & $(-\infty, \infty)$ & conv & 2 & 1x1 & 1,1 & - \\
$d$               & $[0, d_{max}]$ & conv & 1 & 1x1 & 1,1 & $d_{max} \cdot \sigma(\cdot)$ \\
$\sin(\theta), \cos(\theta)$ & $[-1, 1]$ & conv & 2 & 1x1 & 1,1 & $\tanh(\cdot)$ \\
$w,h,l$           & $(-\infty, \infty)$ & conv & 3 & 1x1 & 1,1 & - \\
\hline
\end{tabular}
\vspace{4pt}
\caption{The 3D detector has 5 decoders with differing value ranges, number of filters and activation functions. Each decoder consists of a convolution layer with the parameters given in the table. The bounding box is encoded as the center $c_x, c_y$ in image coordinates relative to the output position. The distance $d$ from the camera, the orientation $\theta$ encoded as $\sin(\theta), cos(\theta)$ - to avoid inconsistencies at $0$ and $2\pi$ - and the dimensions of the bounding box width $w$, height $h$ and length $l$.}
\label{tab:decoders}
\end{table*}

The original focal loss and the automated focal loss can also be applied to regression.
We propose a new method to convert regression predictions into probabilities that can be used for focal loss.
Beyond usage in focal loss, these probabilities will also give a representation of the training progress that is independent of value ranges of the problem.

The core idea to apply focal loss to regression problems is to compute the probability that a prediction is better than the labels.
The underlying assumption is that labels are not perfectly accurate and have an error.
Using the deviation between prediction and ground truth $\Delta x$ and  the deviation between the label and the ground truth $\Delta x_t$ leads to the formulation of $p_{correct}$ as the following probability that $\Delta x_t$ is not between $\pm |\Delta x|$.
\begin{equation}
    p_{correct}
    = 1 - p(-|\Delta x| < \Delta x_t < |\Delta x|)
\end{equation}
We assume that the labels are distributed around the actual correct ground truth by a Gaussian distribution with a variance of $\sigma^2$.
With this assumption it is possible to compute the probability  $p_{correct}$ using the cumulative distribution function $\Phi$ of the Gaussian distribution.
\begin{equation}
\label{eq:regression_p}
    p_{correct}
    = 1 - (\Phi(\frac{|\Delta x|}{\sigma^2}) - \Phi(-\frac{|\Delta x|}{\sigma^2}))
\end{equation}
However, to correctly compute the cumulative distribution function the variance $\sigma^2$ of the task needs to be estimated.
As elaborated in detail in~\cite{kendall17} the uncertainty related to a task which is inherent to the task and the labels can be estimated by adding $\log(\sigma^2 + 1)$ to the loss and training the variable $\sigma^2$ like a weight of the network.
This leads to the final form of the focused loss for regression using $p_{correct}$ as defined in equation~\ref{eq:regression_p} with a $\gamma$ that can be either be constant as in the classical focal loss or even automated as proposed in equations~\ref{eq:focal_loss_gamma_via_integral_h} and~\ref{eq:focal_loss_gamma_via_log}:
\begin{equation}
    L_{focused} = w \cdot L + \log(\sigma^2 + 1)
\end{equation}
with
\begin{equation}
    w = (1-p_{correct})^\gamma
\end{equation}
This formulation leads to an implicit trade-off shift during training.
At the beginning of the training, the loss can focus on examples with a low probability of being correct, like for the automated focal loss on classification.
With progression in the training as the probabilities increase the focus shifts and examples with a low probability are no longer weighted more than examples with a medium probability as seen in Figure~\ref{img:info_vs_h}.
This leads to a reduced impact of outliers on the performance of the network.

The beauty of the formulation as a probability is beyond the capability of computing a focal loss on a regression problem.
The value of $\hat{p}_{correct}$ is independent of the actual value range of the problem,
due to normalizing the absolute distance by the variance of the task $\sigma^2$ - assumed $\sigma^2$ is estimated correctly.
This leads to $\hat{p}_{correct}$ being a new metric on monitoring your training progress.

\section{Detection Network}

For comparability with the original focal loss we chose RetinaNet as the network for predicting detections on the COCO dataset.
Since we propose a new loss the base network remains unchanged from~\cite{lin18}.

However, beyond applying the loss on 2D detection on COCO we applied our newly introduced loss on the very challenging task of 3D vehicle detection.
To keep the focus on the loss and not the network architecture, we decided to use a simple VGG16~\cite{simonyan14} encoder with a decoder specialized on 3D detection as presented in~\cite{weber19}.
The VGG16 encoder consists of the VGG16 network weights pre-trained on the ImageNet dataset~\cite{deng09} up to the layer \textit{pool5}.
Then two 1x1 convolution layers with 4096 filters similar to the fully connected layers from the original VGG network are appended.
On the resulting feature maps a 3D decoder is applied.
The 3D decoder consists of 5 parallel streams:
\begin{enumerate}
    \item A 1x1 convolution softmax prediction for the classification $p_{class}$,
    \item a 1x1 convolution layer with 2 filters to predict the center of the object $c_x, c_y$,
    \item a 1x1 convolution layer with 1 filter to predict the distance of the object from the camera,
    \item a 1x1 convolution layer with 2 filters and $\tanh$ activation function to predict the orientation $\theta$ encoded as $\sin(\theta)$ and $\cos(\theta)$,
    \item a 1x1 convolution layer with 3 filters to predict the size of the object encoded as width, height and length ($w, h, l$) of the object.
\end{enumerate}

All details on the value ranges and the number of filters can be found in Table~\ref{tab:decoders}.

\section{Experiments}

We present experimental results on the bounding box detection task of the COCO benchmark~\cite{lin14} and the challenging KITTI 3D Object Detection dataset~\cite{geiger12}.
We will not focus on the architecture of the network but rather comparing results achieved with different losses.

On COCO we evaluated our loss by training with the same parameters as the original ResNet50~\cite{he16} based RetinaNet in~\cite{lin18}.
On our hardware the original focal loss implementation achieved an AP of $30.41$ slightly lower than the $30.5$ we expected from the paper however this is plausible due to differences in the hardware setup and random initialization.
The proposed automated focal loss achieved an AP of $30.38$ without $\alpha$-balancing. Note that the original focal loss can only achieve $30.41$ with $\alpha$-balancing.
The automated focal loss with focal regression converged after only 30 hours, whereas the original focal loss with a constant $\gamma$ converged after 44 hours.
This means the automated focal loss with focal regression converged at approximately $\frac{2}{3}$ the time the original focal loss required.

We found that during large parts of the training the value for $\gamma$ computed as the information in a correct prediction $I(correct)$ stayed at a value of $\gamma = 2.2$ which is very close to the value that~\cite{lin18} found at $\gamma = 2$.
Only at the beginning of the training $\gamma$ started out at a value larger than $6$ converging slowly towards $0$ once the AP reached a plateau at around 30.0.
This demonstrates that automated focal loss is capable of finding the optimal $\gamma$ for a given problem in only one training run and converging towards a similar result as the focal loss with a $\gamma$ that was hand tuned with several experiments.

When inspecting the AP$_{50}$ we found automated focal loss with focal regression to have a far superior AP$_{50}$ of $51.18$ compared to the AP$_{50} = 46.58$ reproduced with the original focal loss.
Using our loss yields an improvement of $4.6$ AP$_{50}$ while keeping the AP the same. An overview of the COCO results can be found in Table~\ref{tb:coco}.

\begin{table*}[t]
\centering
COCO benchmark\\
\vspace{4pt}
\begin{tabular}{l*{6}{c}r}
\hline
Approach          & AP & AP$_{50}$ & Time to Converge \\
\hline
ResNet 50 - 400 \cite{lin18}      & 30.5 & 47.8 & N.A. \\
Reproduced \cite{lin18} & \textbf{30.41} & 46.58 & 44 h \\
Ours (automated + focal regression) & 30.38 & \textbf{51.18} & \textbf{30 h} \\
\hline
\end{tabular}
\vspace{4pt}
\caption{The values stated in \cite{lin18} are slightly higher than those we were able to reproduce. Our automated focal loss using focal regression performs the same as a normal focal loss with $\alpha$-balancing when comparing the AP. However, the AP$_{50}$ is significantly improved when using the our loss. Furthermore the time required until the network converges is significantly reduced with our loss since it is capable to adapt to the training progress and especially at the beginning makes quicker progress than a focal loss with constant $\gamma$.}
\label{tb:coco}
\end{table*}

Beyond evaluating on COCO and comparing to focal loss we tested our approach on the challenging task of 3D detection.
We used the VGG16 Encoder with a 3D decoder.
The dataset consists of 7481 images posing the challenge to learn a difficult task on few data.
We evaluated the simple network trained with a normal loss, $\alpha$-balanced loss, multiloss~\cite{kendall17}, our automated focal classification loss and finally our automated focal classification and regression loss.
The input image size was a random crop of size 256~x~256 pixels which contained at least one detection.
Each image of the original dataset was cropped in 20 different ways and augmented with geometric augmentation such as horizontal flipping and micro-translations as well as texture augmentation such as intensity, contrast, saturation modification and color jitter (as introduced in~\cite{romera18}).

For training the networks we used the same hyperparameters for all losses.
As optimizer we used Adam~\cite{kingma15} with default parameters.
For the normal loss and the $\alpha$-balanced loss we chose the weights for all regression losses to be 1.0 and 10.0 for the cross entropy classification loss.
The value of $\alpha$ was computed for every batch as $1-h$ where $h$ is the relative frequency of the class.
The learning rate was exponentially decayed starting at 0.0001 and ending at 0.000001 after 160,000 training steps with a batch size of 16.
Training was done in 24 hours on a single NVIDIA GTX~1080~Ti with Tensorflow 1.10.

Since we have already shown on COCO that our loss is capable of finding an optimal $\gamma$ we decided to not test focal loss on 3D detection, since the comparison would be unfair, whereas there is no prior work on what $\gamma$ is optimal.
The automated focal classification loss outperformed the normal loss, $\alpha$-balanced loss and the multiloss on AOS with $37.0$ compared to the next best result at $36.1$ achieved by multiloss~\cite{kendall17}.
Adding automated focal regression loss increased the AOS by $0.3$.
The full report of all AOS scores and AP scores can be found in Table~\ref{tb:kitti3d}.

\begin{table*}[t]
\centering
KITTI 3D Object Detection\\
\vspace{4pt}
\begin{tabular}{l*{6}{c}r}
\hline
Loss function      & top down AP & AOS & prediction FPS \\
\hline
Normal                                        & - & 35.5 & 28.40 \\
$\alpha$-balance                              & - & 35.9 & 28.41 \\
Multiloss                                     & 20.1 & 36.1 & 28.34 \\
Ours (automated focal classification)              & 24.5 & 37.0 & \textbf{28.43} \\
Ours (automated focal classification + regression) & \textbf{25.0} & \textbf{37.3} & 28.18 \\
\hline
\end{tabular}
\vspace{4pt}
\caption{On the KITTI 3D Object Detection dataset where no optimal $\gamma$ is known, we applied automated focal loss and automated focal loss with regression. This eliminates the need to empirically determine $\gamma$. The results of our loss function are superior to using just a normal loss, $\alpha$-balancing or multiloss. Our loss achieves these values without the need for $\alpha$-balancing, hard negative mining or hyperparameter tuning. On the AP from the top down view of the scene we significantly outperform multiloss whilst also outperforming on the average orientation score (AOS). All losses trained the same network designed for simplicity and prediction speed.}
\label{tb:kitti3d}
\end{table*}

\section{Conclusion}

Focal loss reduces the impact of class imbalance but still relies on $\alpha$-balancing and picking a convenient focal factor $\gamma$ for the task.
Having a constant $\gamma$ it is unable to adapt its focus to the current training progress.
We presented an approach to overcome this by computing $\gamma$ dependent on the training progress and introducing a new kind of loss that is capable to shift its focus during training.
Our experiments on COCO showed that even though a good constant $\gamma$ can achieve the same AP as our loss, we outperform regarding AP$_{50}$ and time to convergence in the training process without the need of any hyperparameter tuning to achieve this.
We further introduced a novel technique to compute a probability for a regression loss to achieve better performance than the task dependent variance.
This enables us to introduce a focal regression loss and a new metric to monitor the training progress of a regression task independent of the value range.
We showed that on the KITTI 3D Object Detection dataset our automatic focal loss outperformed other losses.

{\small
\bibliographystyle{ieee}
\bibliography{egbib}
}

\end{document}